\documentclass{article}

\usepackage[english]{babel}

\usepackage[letterpaper,top=2cm,bottom=2cm,left=3cm,right=3cm,marginparwidth=1.75cm]{geometry}

\usepackage{amsmath}
\usepackage{graphicx}
\usepackage[colorlinks=true, allcolors=blue]{hyperref}

\usepackage{fancyvrb}
\usepackage{listings}
\usepackage{authblk}
\usepackage{subcaption}

\title{Upcycling Large Language Models into Mixture of Experts}
\author{Ethan He\thanks{equal contribution. Correspondence to: \{yihuih,akhattar,rprenger\}@nvidia.com}, Abhinav Khattar$^*$, Ryan Prenger$^*$, Vijay Korthikanti, Zijie Yan, Tong Liu, Shiqing Fan, Ashwath Aithal, Mohammad Shoeybi, Bryan Catanzaro\\
NVIDIA} %
\date{}

\begin{document}
\maketitle
\begin{abstract}
Upcycling pre-trained dense language models into sparse mixture-of-experts (MoE) models is an efficient approach to increase the model capacity of already trained models. However, optimal techniques for upcycling at scale remain unclear. In this work, we conduct an extensive study of upcycling methods and hyperparameters for billion-parameter scale language models. We propose a novel ``virtual group" initialization scheme and weight scaling approach to enable upcycling into fine-grained MoE architectures. Through ablations, we find that upcycling outperforms continued dense model training. In addition, we show that softmax-then-topK expert routing improves over topK-then-softmax approach and higher granularity MoEs can help improve accuracy. Finally, we upcycled Nemotron-4 15B on 1T tokens and compared it to a continuously trained version of the same model on the same 1T tokens: the continuous trained model achieved 65.3\% MMLU, whereas the upcycled model achieved 67.6\%. Our results offer insights and best practices to effectively leverage upcycling for building MoE language models. Code is available\footnote{\url{https://github.com/NVIDIA/Megatron-LM/tree/0431153bf1b5c405057b158189c260107d8b7c3a/megatron/core/transformer/moe\#upcycling}}.
\end{abstract}

\section{Introduction}

Sparse Mixture of Experts (MoE) models~\cite{shazeer2017outrageously} are becoming increasingly popular~\cite{lepikhin2020gshard,jiang2024mixtral,fedus2022review,zoph2022st,zhou2022mixture,du2022glam} since they can help achieve better accuracy without a commensurate increase in model training compute. Most recently, state-of-the-art LLMs like Grok-1\footnote{https://x.ai/blog/grok-os}, DBRX\footnote{https://www.databricks.com/blog/introducing-dbrx-new-state-art-open-llm}, Phi-3.5\footnote{https://huggingface.co/microsoft/Phi-3.5-MoE-instruct}, Mixtral 8x22B~\cite{jiang2024mixtral}, DeepSeek-V2~\cite{deepseekai2024deepseekv2strongeconomicalefficient} and Qwen2~\cite{yang2024qwen2} are MoE models. However, an immense amount of compute has been spent on pre-training dense LLMs with only one MLP layer (one expert)~\cite{wu2024llama,dubey2024llama,adler2024nemotron,parmar2024nemotron,bi2024deepseek}. These existing dense models may be able to achieve better accuracy for the same compute cost if they had access to more parameters through MoEs.
Upcycling pre-trained dense language models into sparse mixture-of-experts models (referred to as simply `upcycling' in this work) has emerged as an efficient approach to increase model capacity without the need for training from scratch~\cite{komatsuzaki2022sparse,dai2024deepseekmoe,bai2023qwen,jiang2024mixtral}. By leveraging the knowledge captured in existing dense checkpoints, upcycling enables the creation of large-scale mixture-of-experts (MoE) models while reducing the computational cost and time required for training.

Most previous work on upcycling either does not provide details into how their models were upcycled~\cite{jiang2024mixtral}, or provides experiments only at a small scale~\cite{komatsuzaki2022sparse}. We also find that the recommendations in ~\cite{yang2024qwen2} lead to sub-optimal models. To improve general knowledge on upcycling methods, we therefore publish this study of upcycling methods and best practices for billion-parameter scale language models.

In this work, we conduct an extensive study of upcycling techniques and hyperparameters. Our contributions are as follows:
\begin{enumerate} 
    \item We recommend training recipes to consistently upcycle billion-parameter scale LLMs.
    \item We perform a comprehensive study to find the best hyperparameters for upcycling including learning rate, batch size, and load balancing loss.
    \item We propose a novel ``virtual group" initialization scheme to enable upcycling into fine-grained MoE architectures, along with a weight scaling approach which brings 1.5\% better loss to both coarse-grained and fine-grained upcycled MoE models.
    \item We compare softmax-then-topK expert routing with the topK-then-softmax approach.
    \item We assess the benefits of higher granularity MoE models and using higher topK.
\end{enumerate}

 We demonstrate that our upcycling approach produces a better model than continued dense model training, softmax-then-topK routing improves over topK-then-softmax approach, and higher granularity MoEs can help boost model accuracy in certain training scenarios. Finally, we upcycle the Nemotron-4 15B model~\cite{parmar2024nemotron} into MoE on 1T tokens and show that it improves MMLU. To make the comparison fair, we train the Nemotron-4 model for an additional 1T tokens and achieve a 65.3\% MMLU score, whereas our Nemotron-4 model which was upcycled on the same 1T tokens achieves 67.6. This shows that the improvement from upcycling is not just due to the additional tokens the model is trained on, but also due to the MoE architecture.

By sharing our upcycling approach and ablation results, we aim to contribute to the growing body of knowledge on efficient and scalable language model development, enabling researchers and practitioners to build upon our work and further advance the field of large-scale MoE models. We use Megatron-LM\footnote{https://github.com/NVIDIA/Megatron-LM}~\cite{shoeybi2019megatron} to upcycle and train our MoE models.

\section{Methodology}
\subsection{Sparse Mixture of Experts}
In this work, we only investigate MoEs on the MLP layer of the transformer. These layers comprise the majority of compute and treat each token individually, avoiding issues with kv-cache consistency.
A routing layer routes the tokens to a subset of multiple possible MLP layers.
This increases the parameter count and presumably the model capacity without necessarily increasing the amount of compute required (measured in total training FLOPs).

\subsubsection{Upcycling}

Upcycling is the approach of converting a trained dense model into an MoE ~\cite{komatsuzaki2022sparse}. The most obvious way to convert a dense model into an MoE model without losing accuracy is to duplicate the MLP layer weights multiple times and use a randomized router, weighting the output of MLP layers by their probabilities
\begin{equation*}
E_1(x) := E_2(x) ... := E_N(x) := FFN(x)
\end{equation*}
\begin{equation}
\label{eq:moe}
MoE\_activation = \sum_{i=1}^{T} P_i \times E_i(x) 
\end{equation}


where $E_i$ are MLP experts with router probability $P_i$ such that $P_i \in topK$. $x$ is the output from attention, $N$ is the total number of experts in the MoE layer, and $T$ is the number of experts every token is routed to (topK).

\begin{figure}[!ht]
    \centering
    \includegraphics[width=1\linewidth]{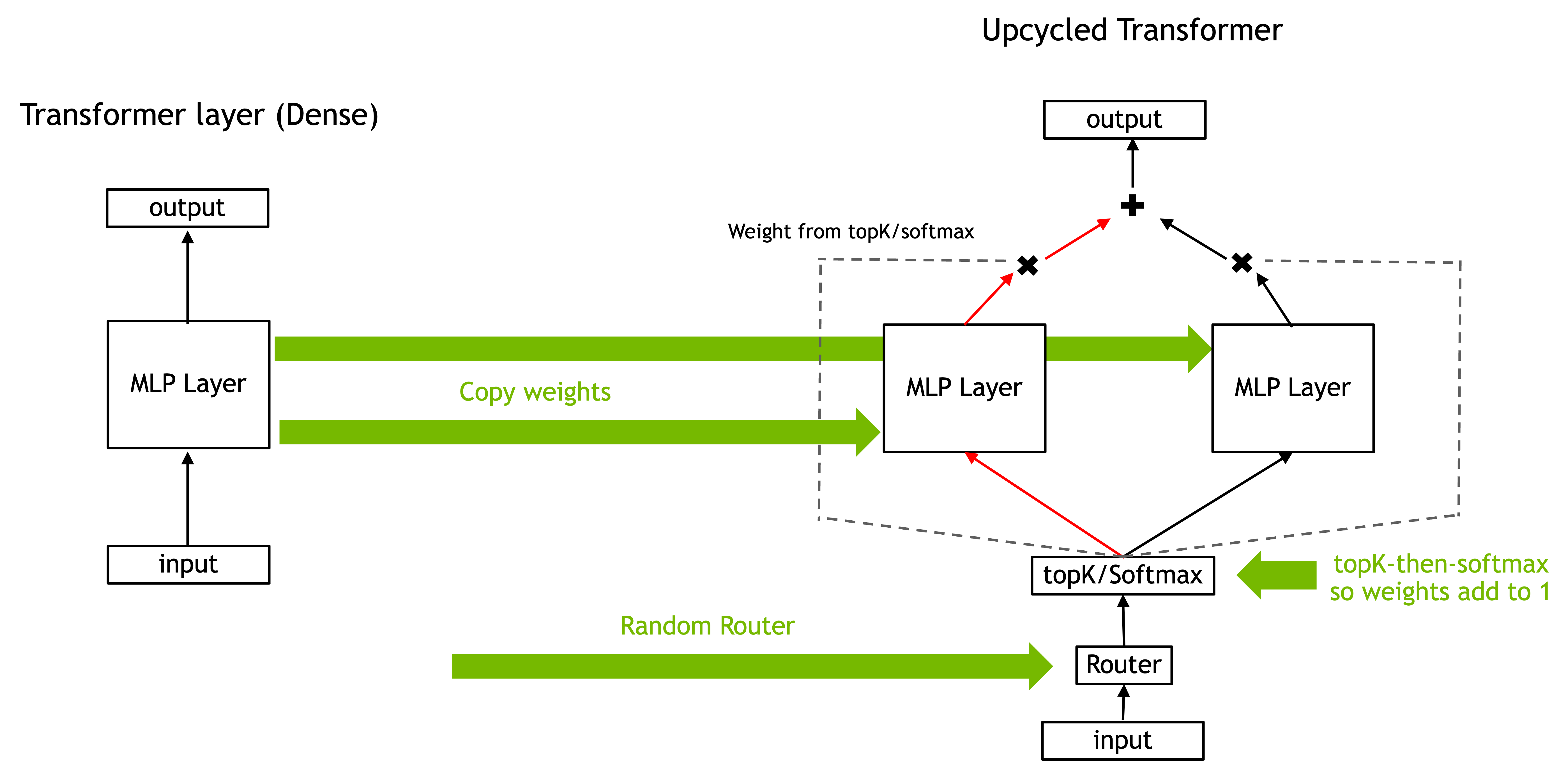}
    \caption{Converting pre-trained dense checkpoint into Mixture of Experts. MLP weights are duplicated to initialize the weights of the experts. The router is randomly initialized. Softmax is often applied after TopK to ensure the upcycled MoE is functionally the same as dense on the first iteration.}
    \label{fig:upcycling_diagram}
\end{figure}

\subsubsection{Softmax - TopK Order}
The standard MoE router formulation~\cite{shazeer2017outrageously,fedus2022switch} performs a softmax on router logits followed by a topK (softmax-then-topK). The activations from the experts are then multiplied by the softmax probability. In this case, the weight of each expert is found using:
\[\text{TopK}(\text{Softmax}(x \cdot W_r))\]
where $x$ is the input to the MoE block and $W_r$ is the router.

This causes an issue with upcycling where the output of the upcycled model is not equivalent to the dense model right after upcycling when TopK $< N$. Even though the output of the upcycled model is not equivalent to the dense model, training the model for a few steps might be able to adjust for this change.

Another way to fix this problem is to use the topK operator directly on the router logits and then only use logits of these topK experts for the softmax computation (topK-then-softmax). In this case, the weight of each expert is found using: 
\[\text{Softmax}(\text{TopK}(x \cdot W_r))\]

The downside of taking this apporach is that the information contained in the absolute magnitude of the router output is lost. Also, this approach only works for topK $> 1$ as softmax of a single element is a constant $1$ which has no gradient w.r.t the input. This technique is used in Mixtral~\cite{jiang2024mixtral}.

In this study, we compare upcycling with both topK-then-softmax and softmax-then-topK in topK$>1$ regime to see which works better (section \ref{sec:st_vs_mixtral}).

\subsection{Granularity}
\begin{figure}[!ht]
    \centering
    \includegraphics[width=1\linewidth]{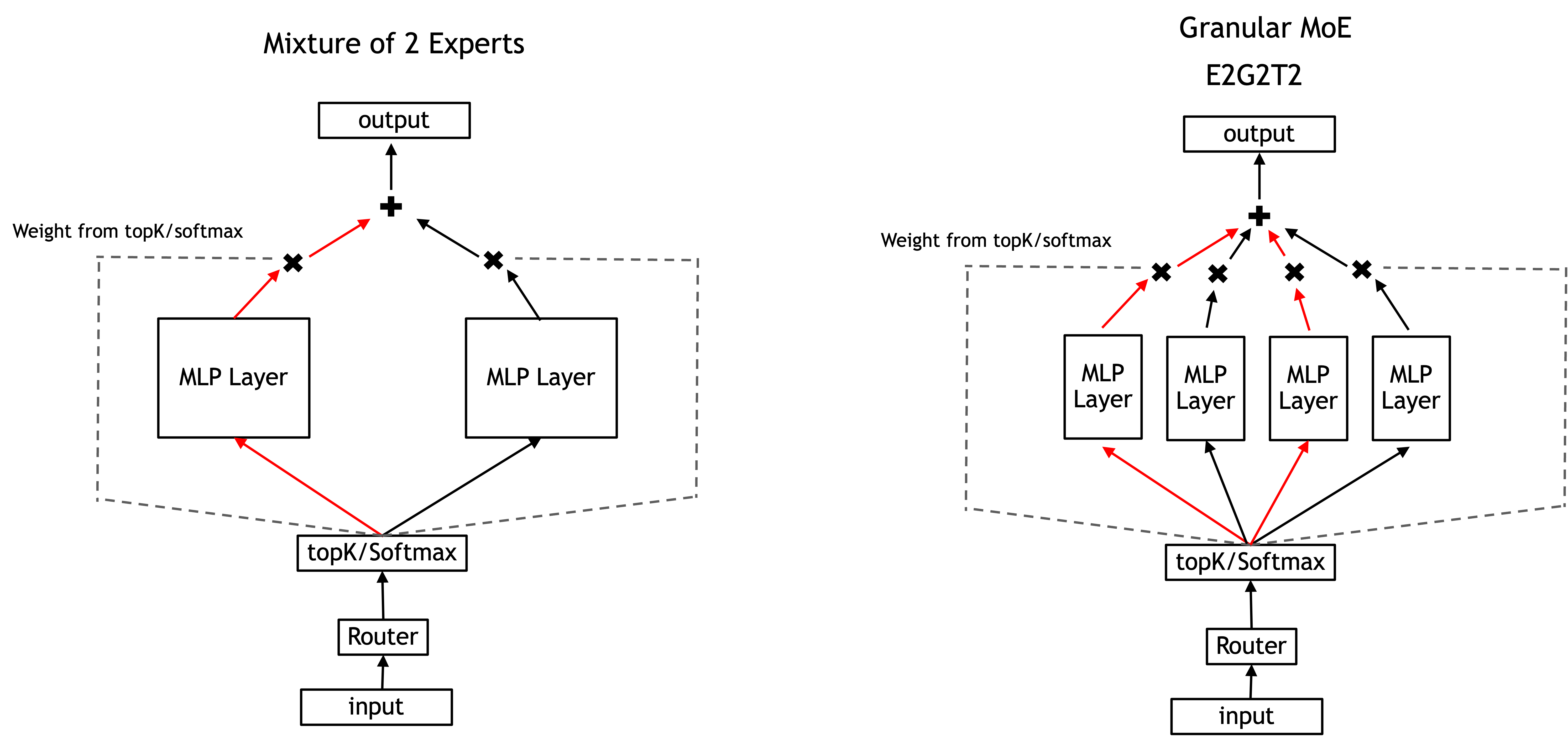}
    \caption{Finegrained Mixture of Experts reduces the size of each expert but activates more experts.}
    \label{fig:granular_diagram}
\end{figure}

Earlier work on MoE routed each token to a very small number of experts (topK $=1$ or $2$)~\cite{fedus2022switch,zoph2022st}.
Routing to only one expert guarantees that the training FLOPs stay similar to the dense model, even though the MoE has more parameters.
However, it has recently been suggested that increasing the number of experts to which a token is routed to, while shrinking the dimension of each expert might be a superior approach \cite{krajewski2024scaling}.
This approach is referred to as granular mixture of experts, shown in Figure~\ref{fig:granular_diagram}.

Granularity introduces a new degree of freedom as every expert can be reduced in size. Since shrinking experts reduces FLOPs per expert, this approach allows us to increase topK by the same magnitude as the shrinking and still keep the overall FLOPs count the same. While FLOPs is only a proxy for the actual compute required to train or deploy a model, it is still useful and an easy metric to compare compute cost.

There are three hyperparameters that define a fine-grained MoE architecture. We use the nomenclature proposed in~\cite{krajewski2024scaling} and add another term $T$ to refer to topK. The three hyperparameters convey the following:
\begin{itemize}
    \item \textbf{$E$: Expansion rate}. How many times larger is the total number of parameters in the MoE layer as compared to the dense MLP counter part. ($N_{\text{MoE}} / N_{\text{dense\_MLP}}$)
    \item \textbf{$G$: Granularity}. How many times smaller is the expert hidden size compared with the original dense layer's FFN hidden size. ($d_{\text{ffn}} / d_{\text{expert}}$)
    \item \textbf{$T$: TopK}. How many experts is a token being routed to.
\end{itemize}
For example, in Figure~\ref{fig:granular_diagram}, from left to right are coarse-grained MoE E2G1T1 and fine-grained MoE E2G2T2.

\subsection{Granular Upcycling} \label{gran_upcycling}
\begin{figure}[!ht]
    \centering
    \includegraphics[width=1\textwidth]{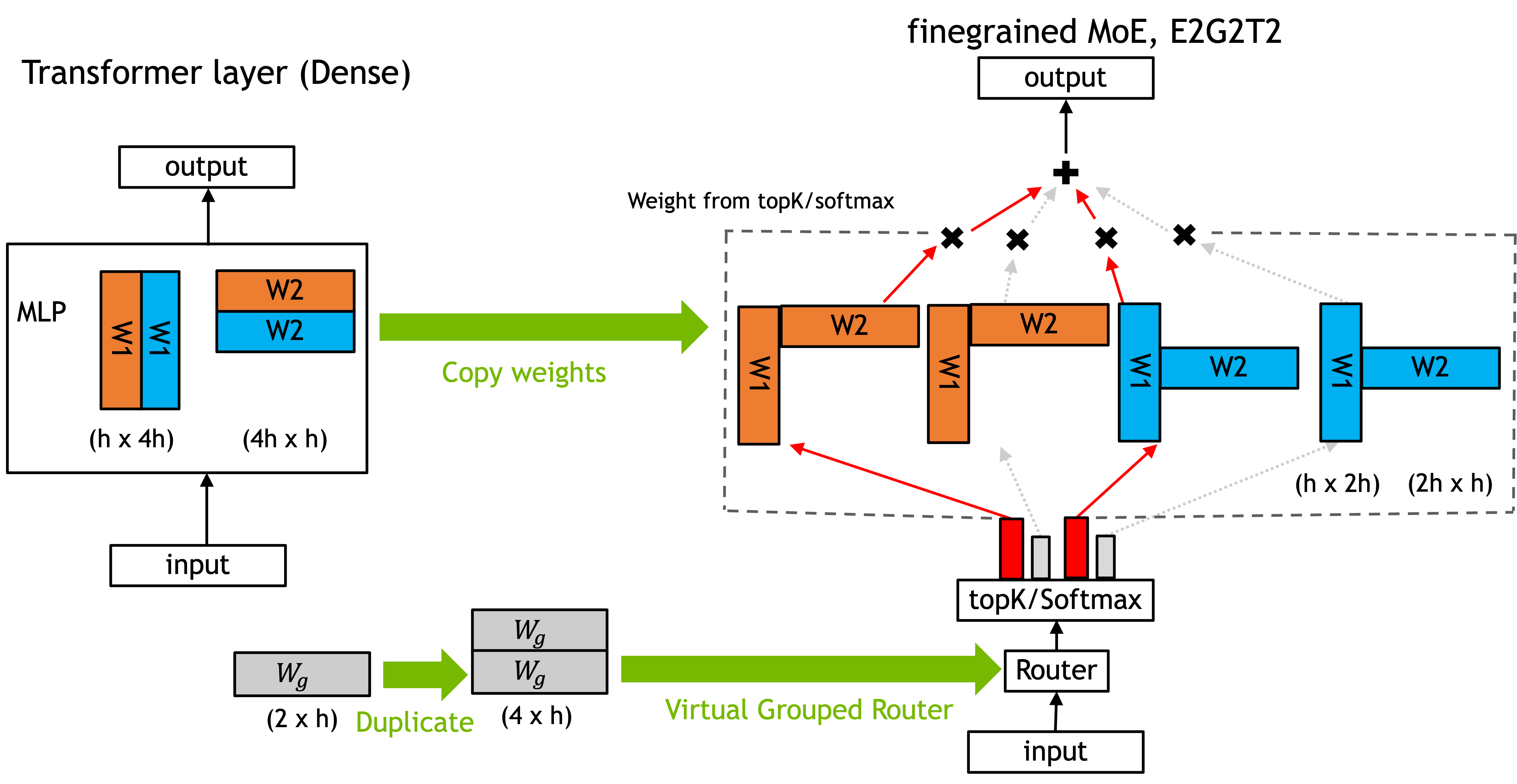}
    \caption{An example of granular upcycling a dense layer into E2G2T2 finegrained MoE. E2G2T2 denotes 4 experts, top 2, with half intermediate size. (1) We shard MLP weights in the intermediate dimension ($4h \rightarrow 2h$) then duplicate the shards. (2) We initialize half the router weights then duplicate them. This ensures Top2 always selects one of each MLP shard so MoE output is the same as the dense model at the start of training.}
    \label{fig:granular_upcycling_diagram}
\end{figure}

Unlike standard MoE upcycling~\cite{komatsuzaki2022sparse} where we can copy the dense MLP weights to MoE experts, granularity reduces the size of every MoE expert. This makes copying the dense MLP weights to MoE experts non-trivial. An intuitive way to upcycle a dense model into finegrained MoE would be to: 

\begin{enumerate}
    \item segment the dense layer into several shards (G) in the FFN dimension.
    \item replicate each shard several times (E). 
    \item route to some experts (T) at training time. 
\end{enumerate}

For example, segmenting into 8 shards then replicating 8 times yields 64 experts. However, using this naive approach we found that the resulting model's loss was very high and the network did not converge to the original loss. There are two problems with this approach: 
\begin{enumerate}
    \item The expert output is scaled down by the router. For 8 expert top-2 model (E8G1T2), a straight forward way of upcycling is to use topK-then-softmax router which ensures that the topK probabilities sum up to 1. However, for fine-grained MoE, even when using topK-then-softmax strategy the outputs are still scaled down. For a 64 experts MoE top-8 model (E8G8T8), if using topK-then-softmax strategy, the output is scaled down by a factor of 8. If using softmax-then-topK strategy, the router probability is about 1/64 because of the random initialization. This implies that the expert outputs are scaled down by a factor of 64.
    \item The MoE is no longer functionally similar to the dense model in the forward pass. For a coarse grained MoE, using the topK-then-softmax strategy, the upcycled model is functionally the same as dense model in the forward pass in the beginning of training. However, for fine-grained MoE, because the experts are segmented into smaller shards, the router needs to select exactly one replica from each segment in order to function the same as a dense MLP layer. 
\end{enumerate}

Motivated by these two observations, we propose weight scaling and virtual grouping for finegrained MoE upcycling. This approach is shown in Figure~\ref{fig:granular_upcycling_diagram}. We initialize the router using \textit{virtual group initialization}. Virtual group initialization ensures that there is exactly one copy of every MLP shard within the router topK right after the dense model is converted into an MoE. Virtual group initialization initializes the experts and router weights such that:
\begin{enumerate}
    \item each router group owns the duplicates of exactly one dense MLP shard.
    \item all router groups have the same router weights.
\end{enumerate}
A pseudo code snippet illustrating this can be found in appendix~\ref{sec:virtual_grouping_code}.

\subsubsection{Scaling the Weights}
We found that with granular upcycling, the scaling of the network weights greatly influences the accuracy of the fine-tuned MoE model. While this scaling could be done entirely in the second linear projection of the MLP ($W2$ in Figure~\ref{fig:granular_upcycling_diagram}), we found empirically that this works worse than scaling both the linear projection weights ($W1$ and $W2$). Equation~\ref{eq:wt_scaling} calculates this scaling factor for the case of squared-relu activation which we use for our base 15B dense model. 

\begin{equation*}
MoE\_activation = \sum_{i=1}^{T} P_i \times E_i(x)
\end{equation*}
where $P_i$ is the probability for the top $i^{th}$ expert, $E_i$ is the corresponding expert layer, $T$ is the topK, and $x$ is the output from attention.\\  \break
assuming approximately uniform distribution\footnote{while this is a simplifying assumption, it holds true for our most important virtual grouping case where the shrinking factor is the same as topK} for iteration 0
\begin{equation*}
P = P_1 = P_2 = ... P_T = \frac{1}{E \times G}
\end{equation*}
\begin{equation*}
MoE\_activation = P \times (\sum_{i=1}^{T} E_i(x)) 
\end{equation*}
So for virtual grouping,
\begin{equation*}
MoE\_activation = \frac{1}{E \times G} (\frac{T}{G} \times dense\_activation) = \frac{T}{E \times G^{2}} \times dense\_activation
\end{equation*}
assuming squared relu activation, we normalize $W_1$ and $W_2$ for each expert's MLP in case of virtual grouping by:
\begin{equation}
\sqrt[3]{\frac{E \times G^{2}}{T}}   
\label{eq:wt_scaling}
\end{equation}

While for different activation functions a hyperparameter search for the optimal scaling of the input and output weights might be intuitively better, we find empirically that the equal weight distribution like above works just as well for our 2B models using swiglu activation. We also use this weight scaling for non-granular (a.k.a coarse-grained) MoE models and observe that it helps convergence.

\section{Results}
\noindent\textbf{Model}: We do all our ablation studies on our Nemotron 2B\footnote{https://huggingface.co/nvidia/GPT-2B-001} and Nemotron-4 15B~\cite{parmar2024nemotron} models followed by final results using a larger token count on Nemotron-4 15B. Nemotron 2B is a transformer-based decoder-only language model similar to GPT-2 and 3~\cite{radford2018improving}. This model was trained on 1.1T tokens using NeMo~\cite{kuchaiev2019nemo} and Megatron-LM~\cite{shoeybi2019megatron}. The model uses the SwiGLU activation function~\cite{shazeer2020gluvariantsimprovetransformer}, Rotary Positional Embeddings (RoPE)~\cite{su2024roformer}, maximum sequence length of 4096, no dropout, no bias terms in all linear layers, and untied embedding and output layer. Nemotron-4 15B is a 15-billion-parameter large multilingual language model trained on 8 trillion text tokens. Both these models use a vocab size of $256K$.

\noindent\textbf{Data}: Upcycling can be performed on either pretraining data that the pretrained dense model has seen, new unseen data, or a combination of both. In our Nemotron 2B experiments, we upcycle on pretraining data the model has seen for simplicity. For all of the ablation studies, we use 110B tokens (about 10\% of the pretraining 1.1T tokens). For Nemotron-4 15B experiments, we upcycle on continued training data so that we can compare upcycling against our existing dense continued training result ~\cite{parmar2024nemotron,parmar2024reusedontretrainrecipe}. While the continued training data has still been seen in pretraining, it follows a different data blending distribution. For 15B ablations, we train on 0.1T tokens (10\%) of continued training data blend. For our 15B final results, we train on the full continued training data blend of 1T tokens. Validation loss is measured on 1\% held-out data.

\subsection{Effectiveness of Upcycling}
\begin{figure}[!ht]
    \centering
    \begin{subfigure}[b]{0.48\textwidth}
        \centering
    \includegraphics[width=1\linewidth]{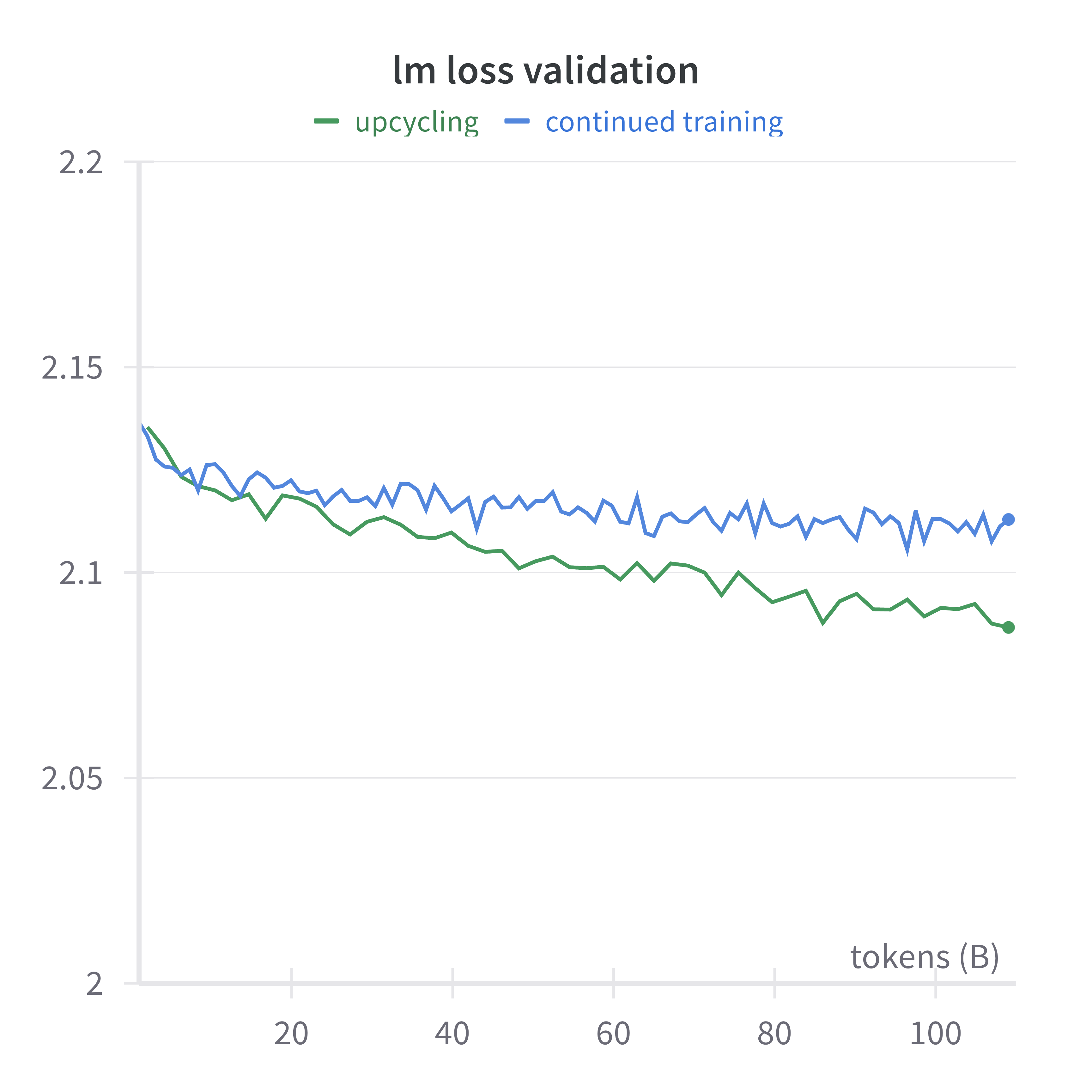}
        \caption{Upcycling Nemotron 2B outperforms continued training on loss}
    \label{fig:upcycling_vs_ct}
    \end{subfigure}
    \hfill
    \begin{subfigure}[b]{0.48\textwidth}  
        \centering 
            \includegraphics[width=1\linewidth]{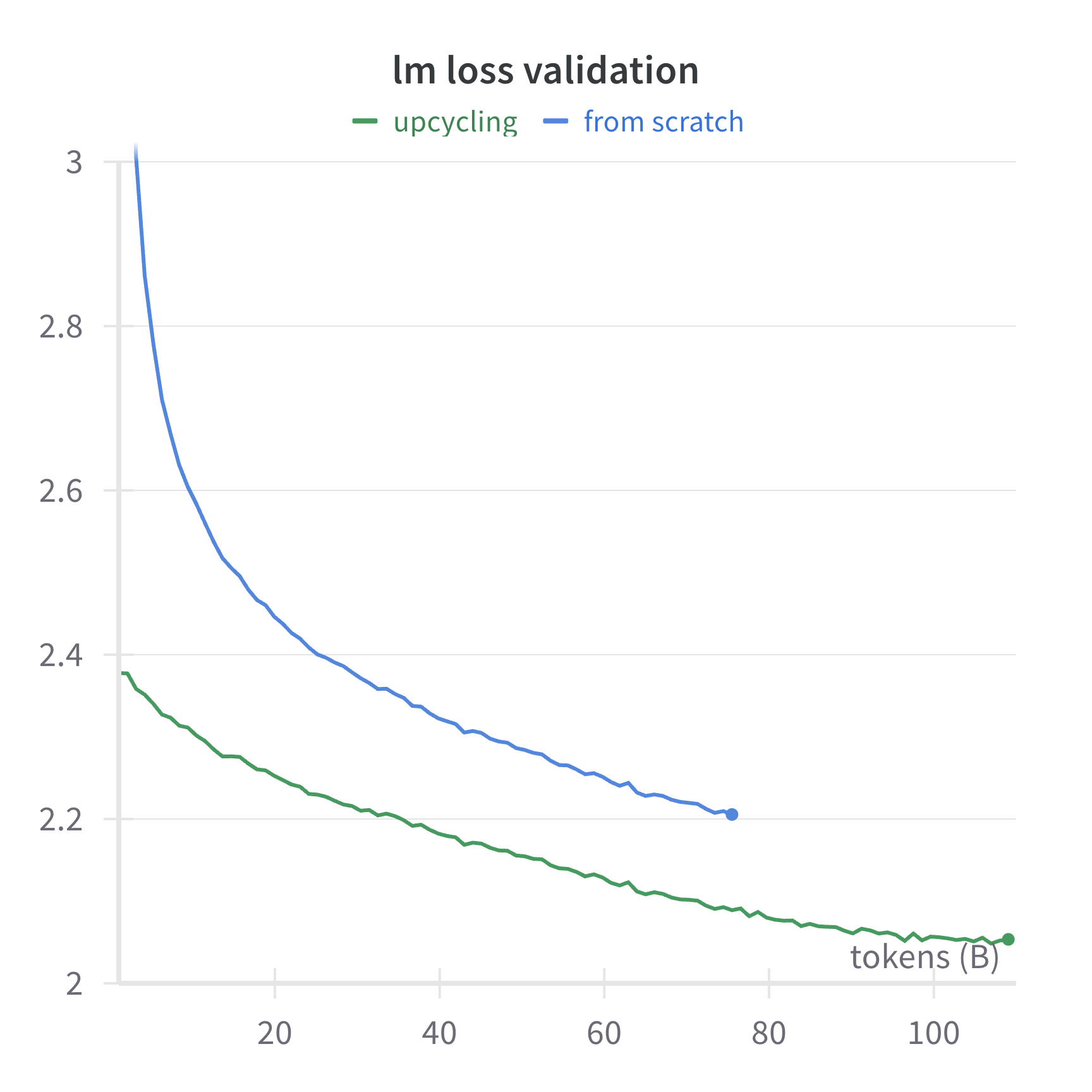}
        \caption{Upcycling Nemotron 2B outperforms training from scratch on loss under certain compute budget}
            \label{fig:vs_scratch}
    \end{subfigure}
    \caption{Effectiveness of upcycling}
\end{figure}

\noindent\textbf{Upcycling vs. Dense Continued Training}: Following previous works~\cite{komatsuzaki2022sparse}, we compare upcycling vs continued training the dense Nemotron 2B model with the same amount of tokens (0.1T) under a similar learning schedule. 
As shown in Figure~\ref{fig:upcycling_vs_ct}, continuous training plateaus quickly while upcycling keeps improving. From continued training to upcycling, LM loss improved by 1.1\%.

\noindent\textbf{Upcycling vs. Training from Scratch}: Figure~\ref{fig:vs_scratch} shows that upcycling achieves good improvement over training from scratch if one assumes a fixed compute budget. Upcycling is an efficient method to utilize pretrained dense model weights when the compute budget is much smaller than the pretraining compute budget. An interesting question that remains unanswered with our studies is whether it is still worth to upcycle a dense model instead of pretraining, assuming a larger compute budget. While some recent works like Skywork-MoE~\cite{wei2024skywork} try to answer this question, we leave investigating this as a potentially interesting future direction.

\subsection{Learning Rate and Batch Size} 
\subsubsection{Learning Rate Schedule}\label{sec:lr}
We found that the learning rate schedule plays an important role in upcycling. When upcycling a dense model, the model has usually been trained for a large number of steps already and the learning rate is generally decayed during this training phase allowing the model to enter a local minimum. For example, our 2B model follows a warmup cosine decay learning rate schedule. The learning rate is warmed up to a 2e-4 peak and then decayed to 2e-5 at the end of 1.1T token training. During upcycling, it is unclear if increasing the learning rate above where pretraining ended (we call this resetting the learning rate) will improve or hurt model quality.

To find a good learning rate schedule for upcycling, we experimented with three different settings:

\begin{itemize}
  \item \textbf{Constant learning rate}. We use the minimum pretraining learning rate of 2e-5 which is a typical learning rate schedule for finetuning. 
  \item \textbf{Peak learning rate 2e-4, cosine decay to 2e-5}. This is the exact same learning rate schedule as pretraining, except that it decays to 2e-5 at the end of 0.1T tokens upcycling, as we only have 10\% of pretraining tokens. 
  \item \textbf{Peak learning rate 1e-4, cosine decay to 2e-5}. Since it's possible that using learning rate as high as pretraining can lead to catastrophic forgetting, using a lower peak learning rate might be a good option.
\end{itemize}

\begin{figure}[!ht]
    \centering
    \begin{subfigure}[b]{0.48\textwidth}
        \centering
        \includegraphics[width=1\linewidth]{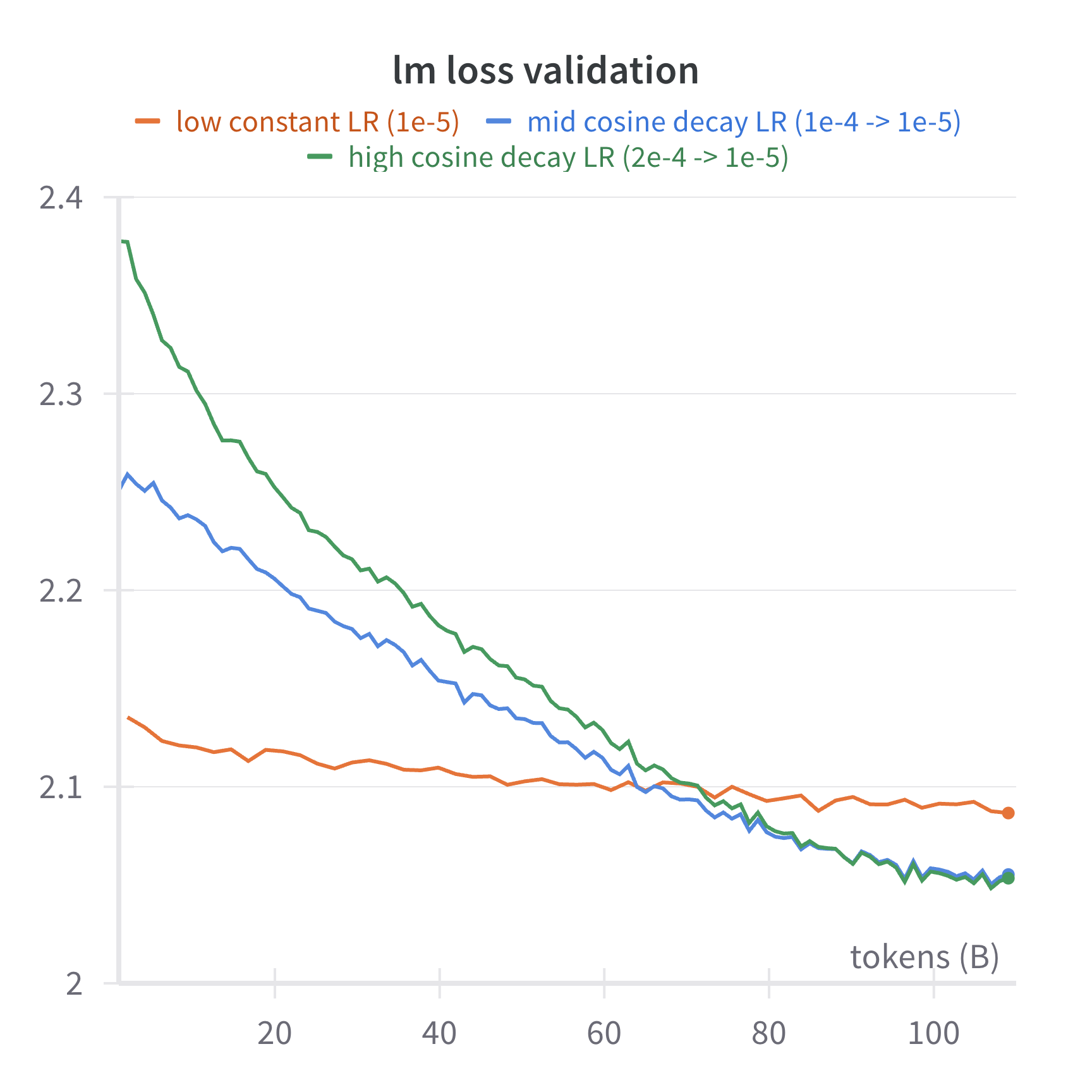}
    \end{subfigure}
    \hfill
    \begin{subfigure}[b]{0.48\textwidth}  
        \centering 
        \includegraphics[width=1\linewidth]{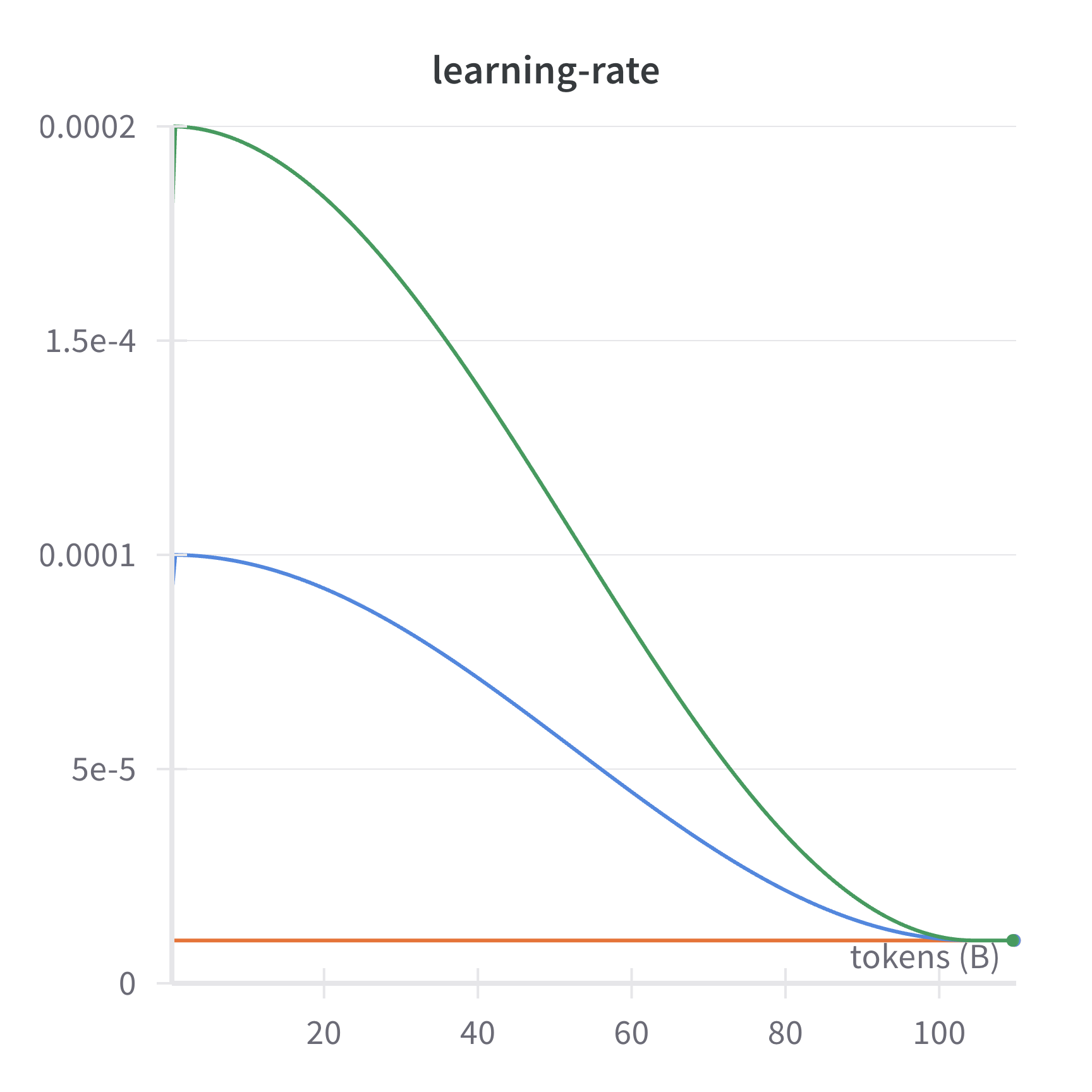}
    \end{subfigure}
    \caption{Resetting LR to peak pre-training LR when upcycling Nemotron 2B improves accuracy.}
    \label{fig:spiking_lr}
\end{figure}

As shown in Figure~\ref{fig:spiking_lr}, we found that while constant learning rate schedule starts off with much lower loss than the reset learning rate schedule, it eventually plateaus. The reset learning schedule gradually catches up and eventually outperforms the constant learning rate schedule. When resetting, peak learning rates of 1e-4 and 2e-4 seem to perform similarly. Interestingly, using a peak learning rate as high as pretraining (2e-4) does not lead to catastrophic forgetting.  

\begin{figure}[!ht]
    \centering
    \includegraphics[width=1\linewidth]{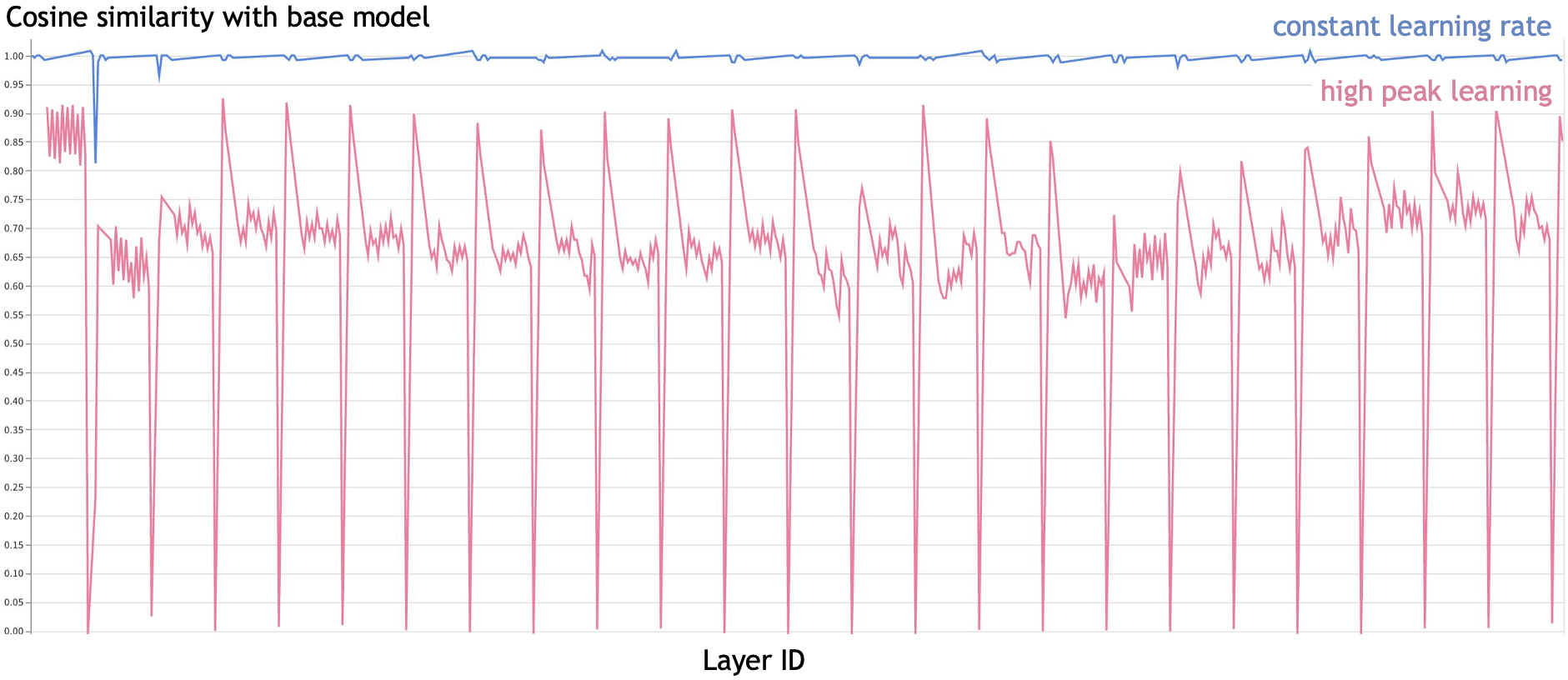}
    \caption{Cosine similarity between upcycled MoE and base Nemotron 2B model weights. When higher learning rate is used, similarity is lower and upcycling validation loss improves. }
    \label{fig:weight_sim}
\end{figure}

\noindent\textbf{Weights Similarity}: Typically, finetuned model weights have very high cosine similarity with the base model weights. For example, Llama 2 chat model has cosine similarity close to 1 with Llama 2. 

We compute cosine simimarity between an upcycled model and the original dense model layer-by-layer and since MoE layer does not match with MLP, we compute cosine similarity between each expert and the orignal MLP and take the average. We finally average the cosine similarity across all layers to get a single number.

Shown in Figure~\ref{fig:weight_sim}, like in most finetuning tasks, upcycling with the minimum learning rate does not change the weights much. The cosine similarity between the MoE and base model is close to 1. By applying the reset learning schedule method, we observed the cosine similarity reduced to 0.6-0.7. This might imply that the higher learning rate helps the model escape from the dense model's local minimum, and find a superior minimum. Additionally, the high learning rate also helps the experts diversify. On the other hand, small constant learning rate leads to experts being very similar to each other, which makes the model not too different from the base dense model.

\subsubsection{Batch Size}
Aside from the learning rate, we observed that batch size also heavily affects MoE training and upcycling. We hypothesize that MoEs benefit from larger batch size than dense equivalents for two reasons: 

\begin{itemize}
  \item Since each expert receives only a portion of tokens, gradients are noisier than the dense model. 
  \item The load balancing loss is noisier if there are fewer tokens to balance.
\end{itemize}

\begin{figure}[!ht]
    \centering
    \begin{minipage}{0.49\textwidth}
        \centering
    \includegraphics[width=1\linewidth]{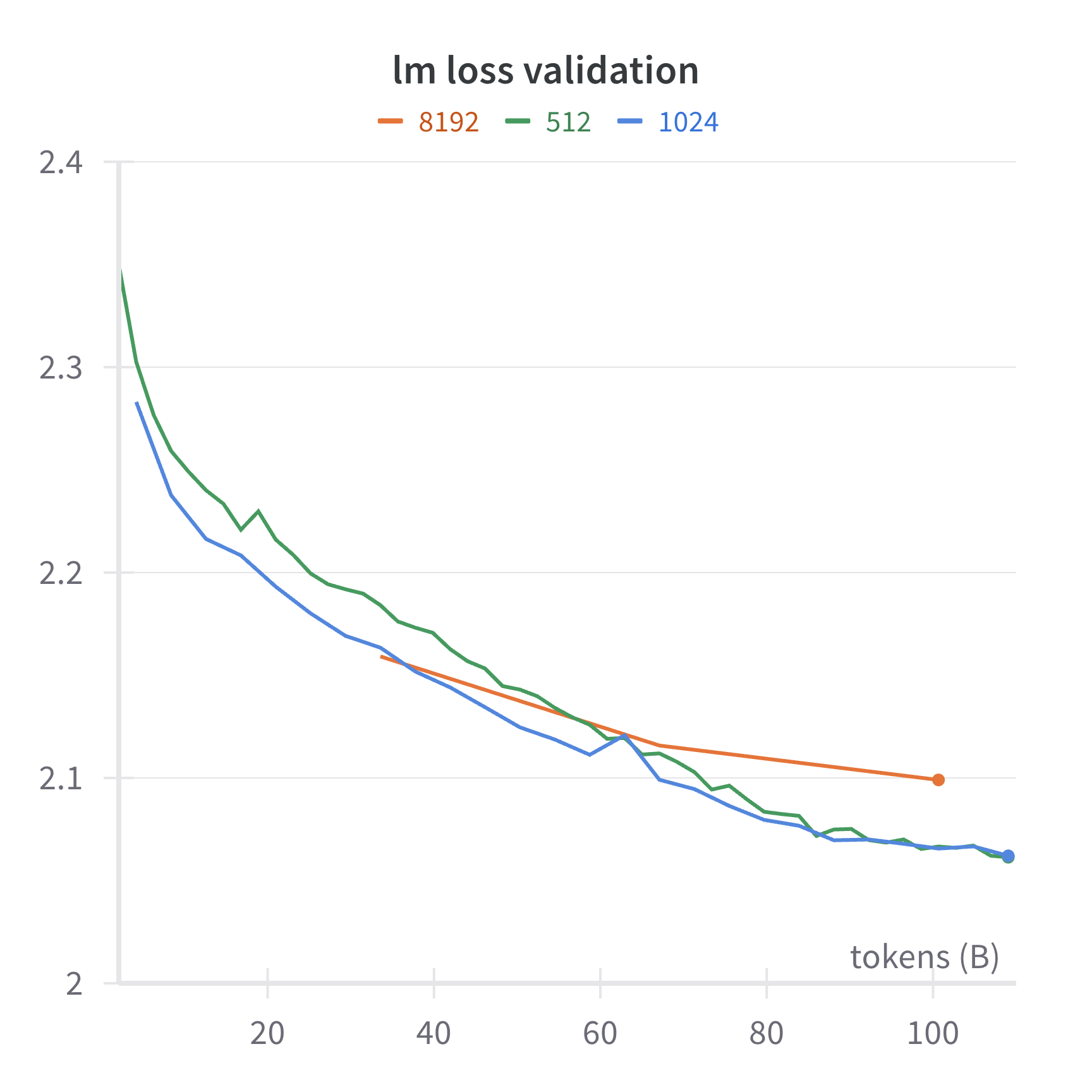}
    \caption{Effect of batch size on validation loss when upcycling Nemotron 2B. Increasing batch size to 1024 (4M tokens) does not degrade accuracy while improving model FLOP utilization. }
    \label{fig:gbs}
    \end{minipage}%
    \hfill
    \begin{minipage}{0.49\textwidth}
        \centering
    \includegraphics[width=1\linewidth]{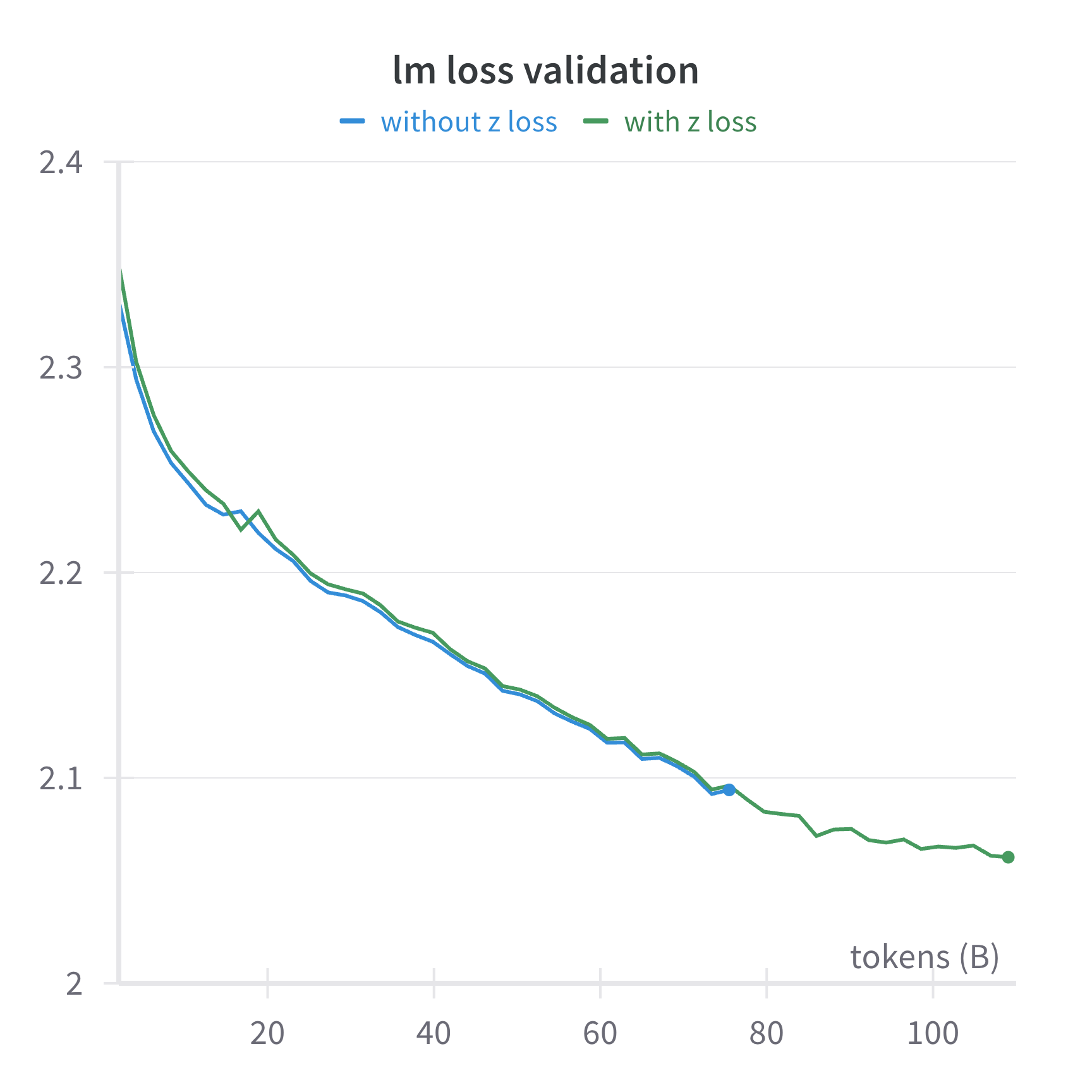}
    \caption{Upcycling Nemotron 2B with or without z loss does not have noticeable difference}
    \label{fig:noz}
    \end{minipage}
\end{figure}

As shown in Figure~\ref{fig:gbs}, we compared batch size of 512, 1024, and 8192 (2M, 4M, and 32M tokens respectively) for upcycling the Nemotron 2B model. While the batch size of 32M tokens performs the worst, batch size of 4M tokens converges faster than 2M tokens. The training efficiency (per-GPU throughput) is also much better with larger batch sizes. 
Recently, Deepseek-V2~\cite{deepseekai2024deepseekv2strongeconomicalefficient} also used a large batch size of 9216 (more than 37M tokens) and learning rate of 2.4e-4 to pretrain MoE models. 

\subsection{Load Balancing and Regularization Loss}
\noindent\textbf{Load Balancing Auxiliary Loss}: We used the same aux loss as described in ST-MoE~\cite{zoph2022st} and Switch Transformer~\cite{fedus2022switch} and experimented with different aux loss coefficients. We found that while not using aux loss at all leads to dead experts and causes the training loss to plateau early, aux loss coefficients set too high leads to aux loss overwhelming the language modeling loss. We found aux loss coefficients between 1e-2 to 1e-3 to give the best language model loss. 

\noindent\textbf{Z Loss}: We used the same z loss described in ST-MoE~\cite{zoph2022st}. As shown in Figure~\ref{fig:noz}, we compared upcycling with and without z loss (with a coefficient of 1e-3) and found that z loss has no impact on the final model quality. 

Thus, we used an aux loss coefficient of 1e-2 and no z loss for all our experiments.

\subsection{Softmax TopK order}\label{sec:st_vs_mixtral}
We consistently found that using softmax-then-topK works better than topK-then-softmax for upcycling. We hypothesize this is because the softmax-then-topK approach preserves the information contained in the absolute value of the router output. However, keeping the output of the original upcycled model similar to the output of the dense model is more difficult with this approach because the outputs no longer sum to one. We overcome this issue with our weight scaling method. 

\subsection{Fixing Output Scale}
We tried multiple approaches to compensate for the issue of expert outputs being scaled down. Apart from weight scaling described in Equation~\ref{eq:wt_scaling}, we also experimented with the following:

\noindent\textbf{Scaling the MoE output}: instead of scaling the weights, we tried scaling the output of the MoE layer by a constant factor or a learnable scalar. We empirically found that neither of them work well. 

\noindent\textbf{Post Expert Layernorm}: Work on finegrained MoE scaling laws~\cite{krajewski2024scaling} recommended adding a layernorm at the end of MoE layer. Typically, dense models do not have this layernorm. We tried adding the post expert layernorm during upcycling and found that while it can achieve the same effect and stop the loss from exploding, it takes a lot longer to adapt.

We compared different methods for upcycling Nemotron 2B into 64 experts top-8 fine-grained MoE (E8G8T8). The expert intermediate hidden size is 1/8 of the original FFN intermediate hidden size so that they are iso-FLOP. Our proposed weight scaling method performed the best and is also the easiest to implement, since it does not modify the model architecture. 

Using equation~\ref{eq:wt_scaling}, for the finegrained MoE with 64 experts top-8, the scaling factor should be 4. We tried multiple scaling factors (2x, 2.5x, 4x) and found that the scaling factor of 4 also performed the best empirically. So our weight scaling function, while not exact, helps convergence.

\begin{figure}[!ht]
    \centering
        \includegraphics[width=0.48\linewidth]{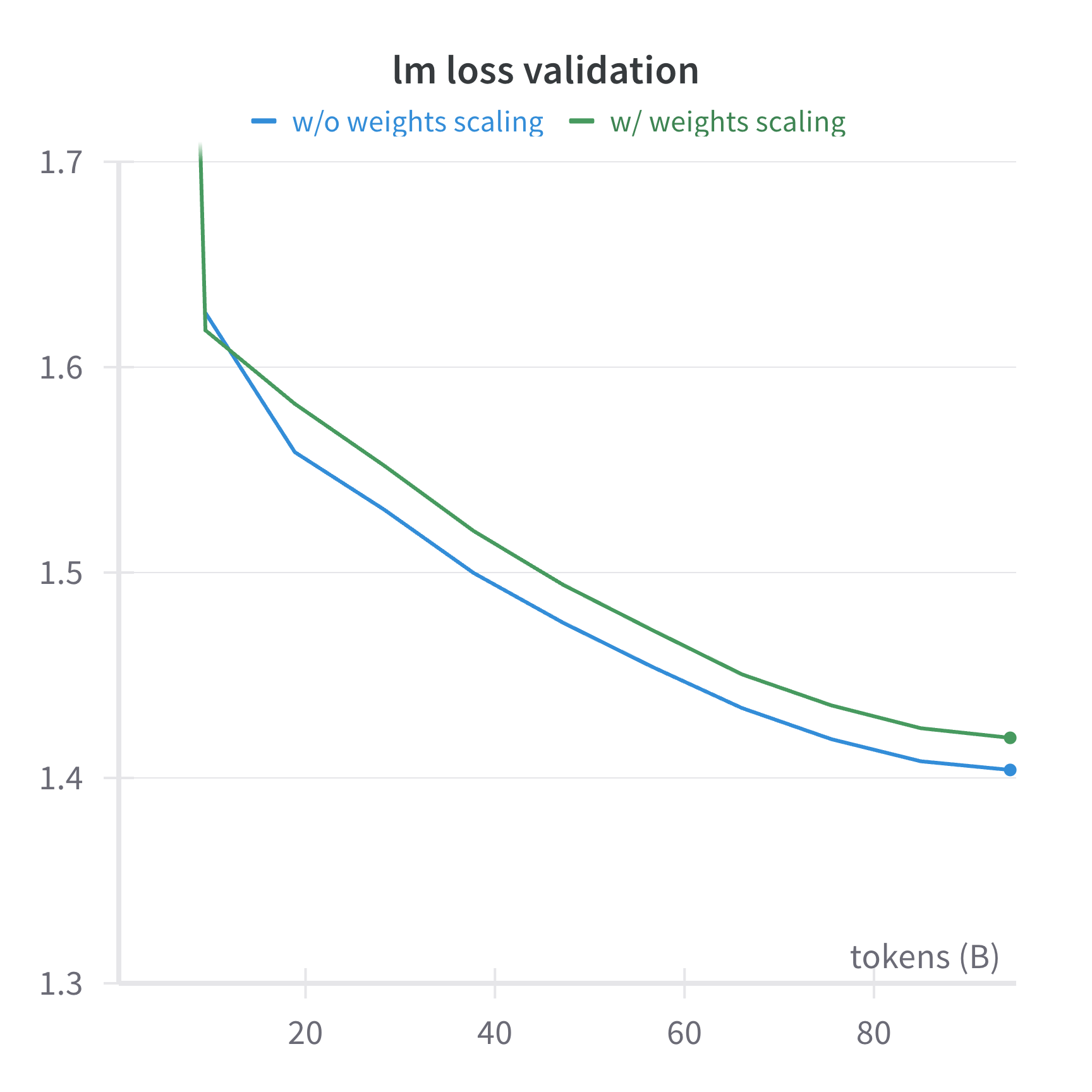}
    \caption{Using weight scaling helps achieve better loss when upcycling Nemotron-4 15B into 8 experts top-1 MoE}
    \label{fig:scaling_15b}
\end{figure}

We also discovered that weight scaling helps upcycling standard coarse-grained MoEs as well. As shown in Figure~\ref{fig:scaling_15b}, we upcycled Nemotron-4 15B into 8 experts top-1. With weight scaling, the training loss is 1.5\% better than when not using weight scaling.

\subsection{Increasing Granularity}
We found that increasing granularity improves loss when training with small token counts (0.1T tokens for our 15B example) since virtual grouped granular upcycling is able to achieve a better loss more quickly than the non-granular version. However, when training on much larger token regimes ($\geq$1T tokens), we saw that granularity did not help proportionally and both granular and non-granular runs converged to similar loss values. Since these large token horizon runs required significant compute, we did not perform ablations on them. 

\begin{figure}[!ht]
    \centering
    \begin{subfigure}[b]{0.48\textwidth}
        \centering
        \includegraphics[width=1\linewidth]{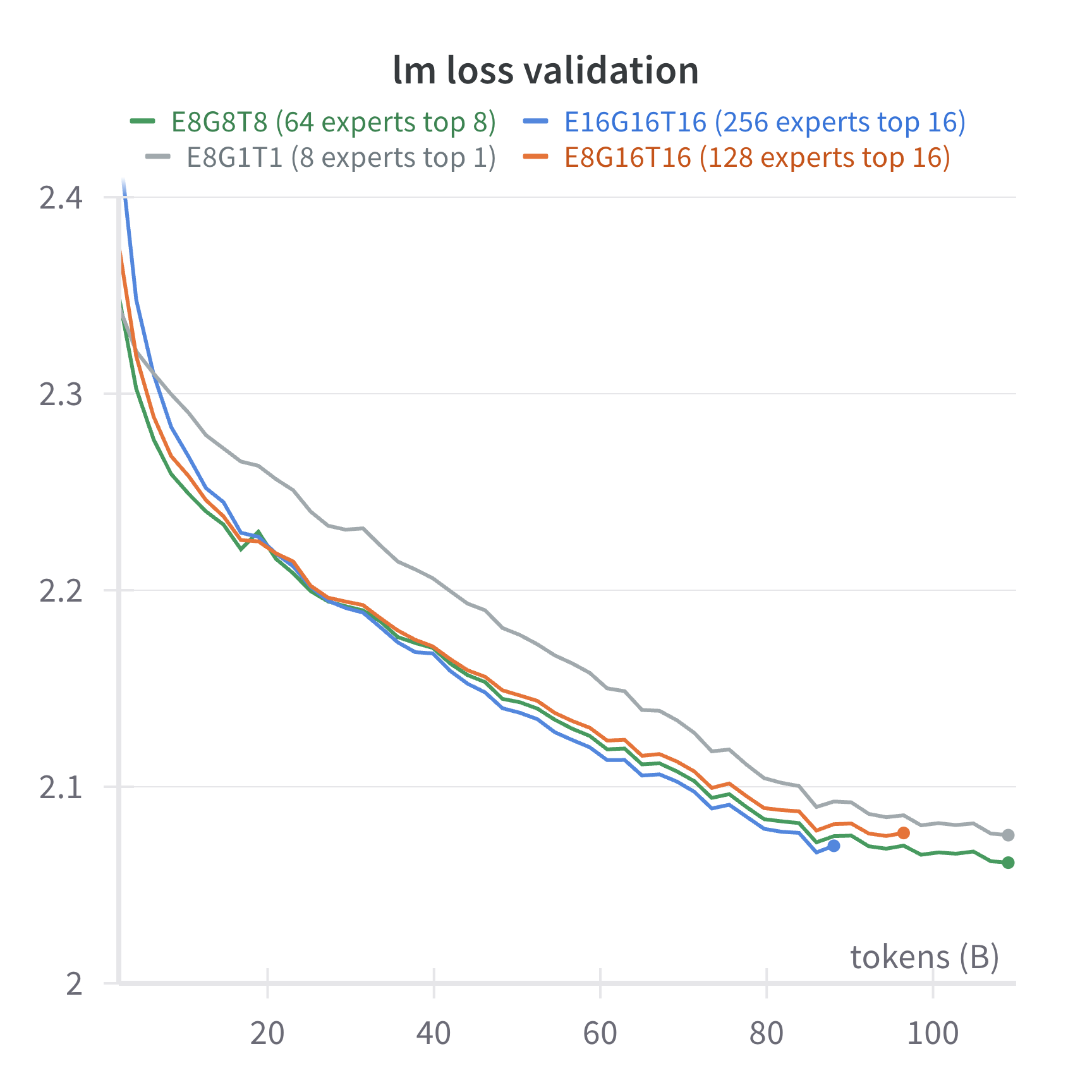}
        \caption{Upcycling iso-FLOP 2B Nemotron}
    \end{subfigure}
    \hfill
    \begin{subfigure}[b]{0.48\textwidth}  
        \centering 
        \includegraphics[width=1\linewidth]{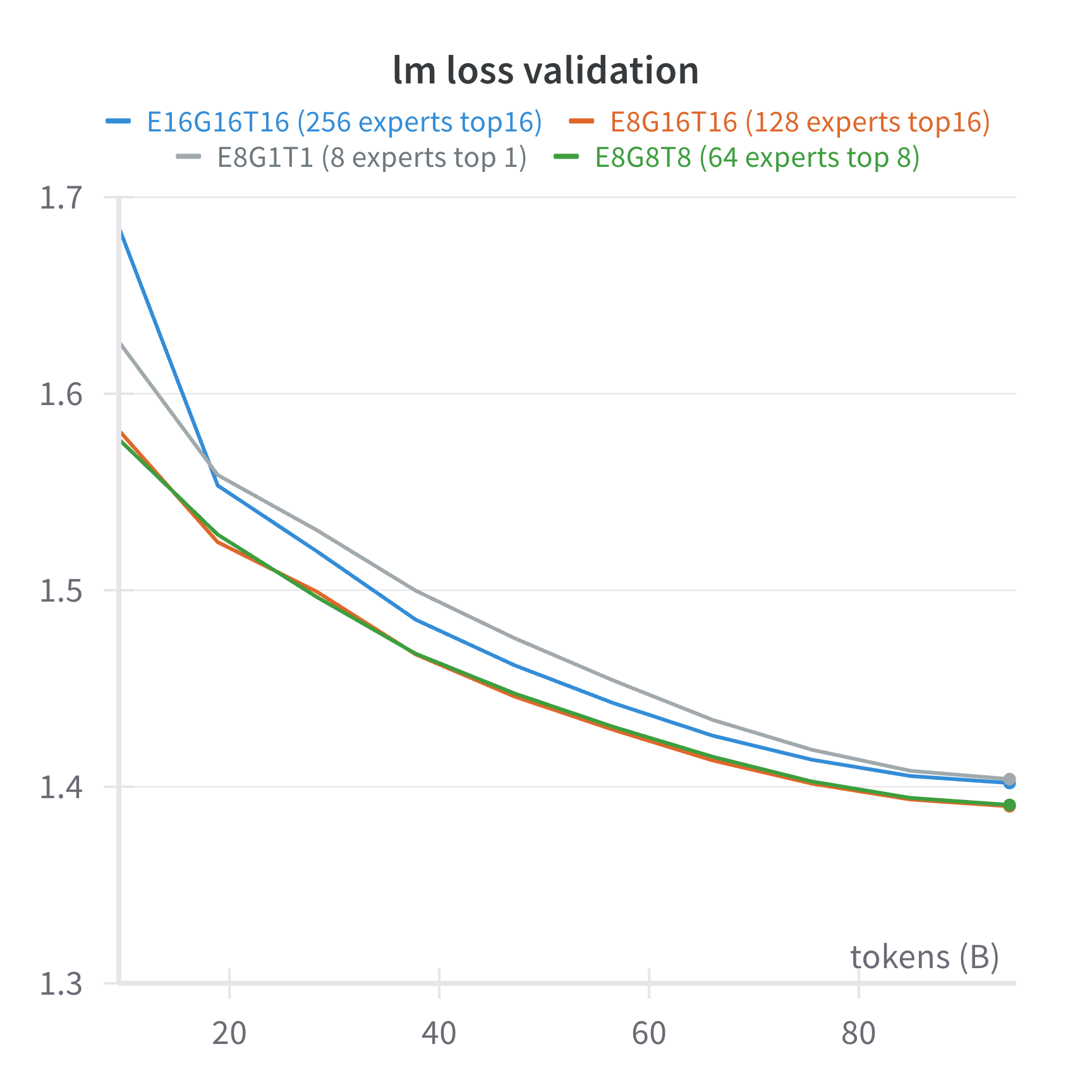}
        \caption{Upcycling iso-FLOP Nemotron-4 15B}
    \end{subfigure}
    \caption{Increasing granularity provides diminishing benefits. We found granularity of 8 to be good for both Nemotron 2B and Nemotron-4 15B.}
    \label{fig:varying_granularity}
\end{figure}

We tried scaling up the number of experts from 8 to 256 without increasing FLOPs. On upcycling Nemotron 2B and Nemotron-4 15B, we compared 8, 64, 128, and 256 experts. We kept all these networks iso-FLOP by scaling down the expert hidden size proportionally with respect to the topK. Shown in Figure~\ref{fig:varying_granularity}, on Nemotron 2B, 64 experts performed better than 8 experts. However, scaling up further to 128 or 256 brought in little benefit. Similarly, on Nemotron-4 15B, the improvement maxed out at 64 experts. Surprisingly, on upcycling Nemotron-4 15B, 256 experts performed slightly worse than 64 or 128 experts. Too many experts seemed to hurt accuracy when upcycling.
We hypothesize that this is because the experts were all copies and the larger the number of experts get, it becomes difficult for the network to find new superior minima. While increasing granularity is promising, it is important to note that it comes with higher MoE permute/unpermute cost and smaller GEMM sizes. Owing to these system-level factors, we worked with both granular and non-granular recipes.

\subsection{Increasing TopK}
\begin{figure}[!ht]
    \centering
    \begin{subfigure}[b]{0.48\textwidth}
        \centering
        \includegraphics[width=1\linewidth]{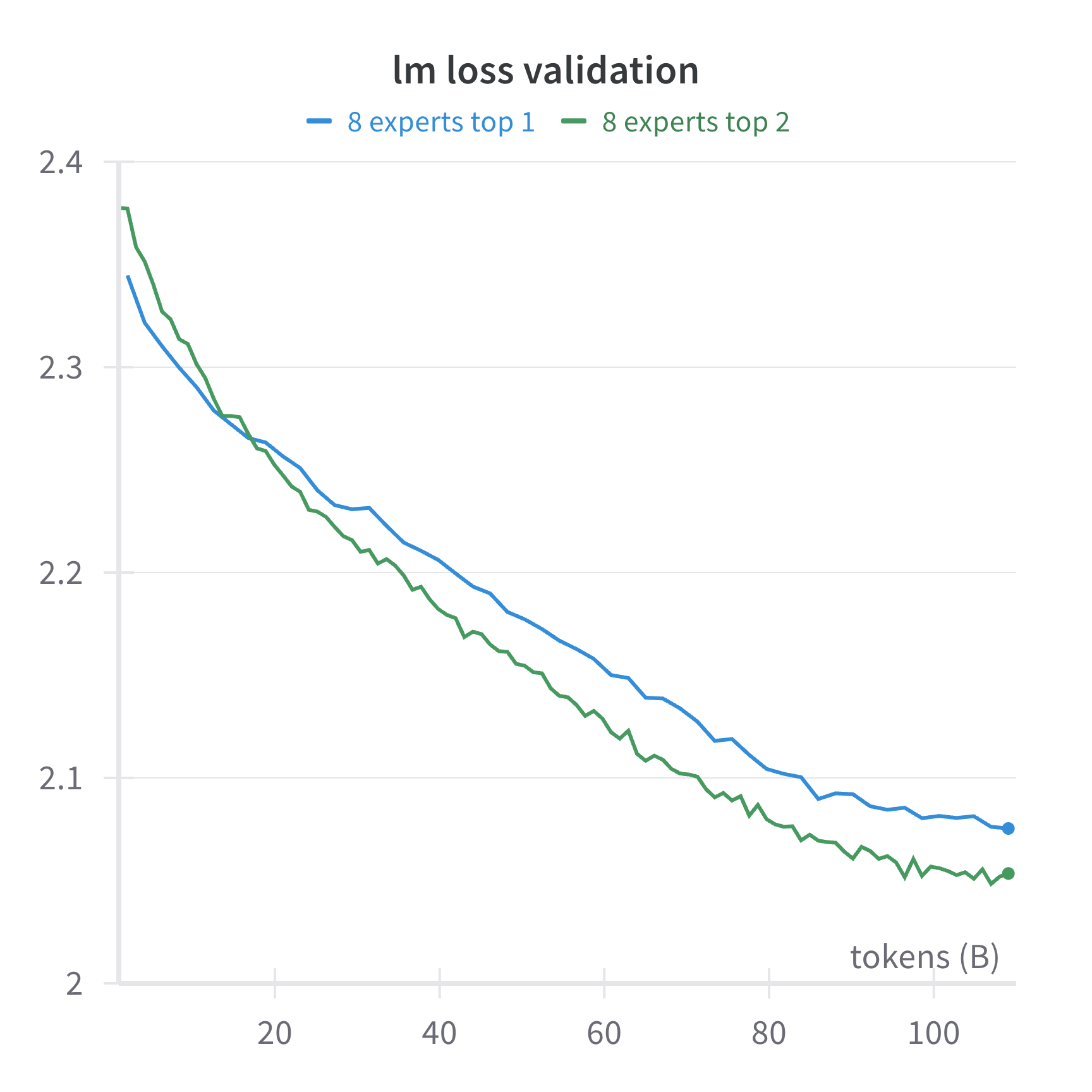}
        \caption{Upcycling Nemotron 2B}
    \end{subfigure}
    \hfill
    \begin{subfigure}[b]{0.48\textwidth}  
        \centering 
        \includegraphics[width=1\linewidth]{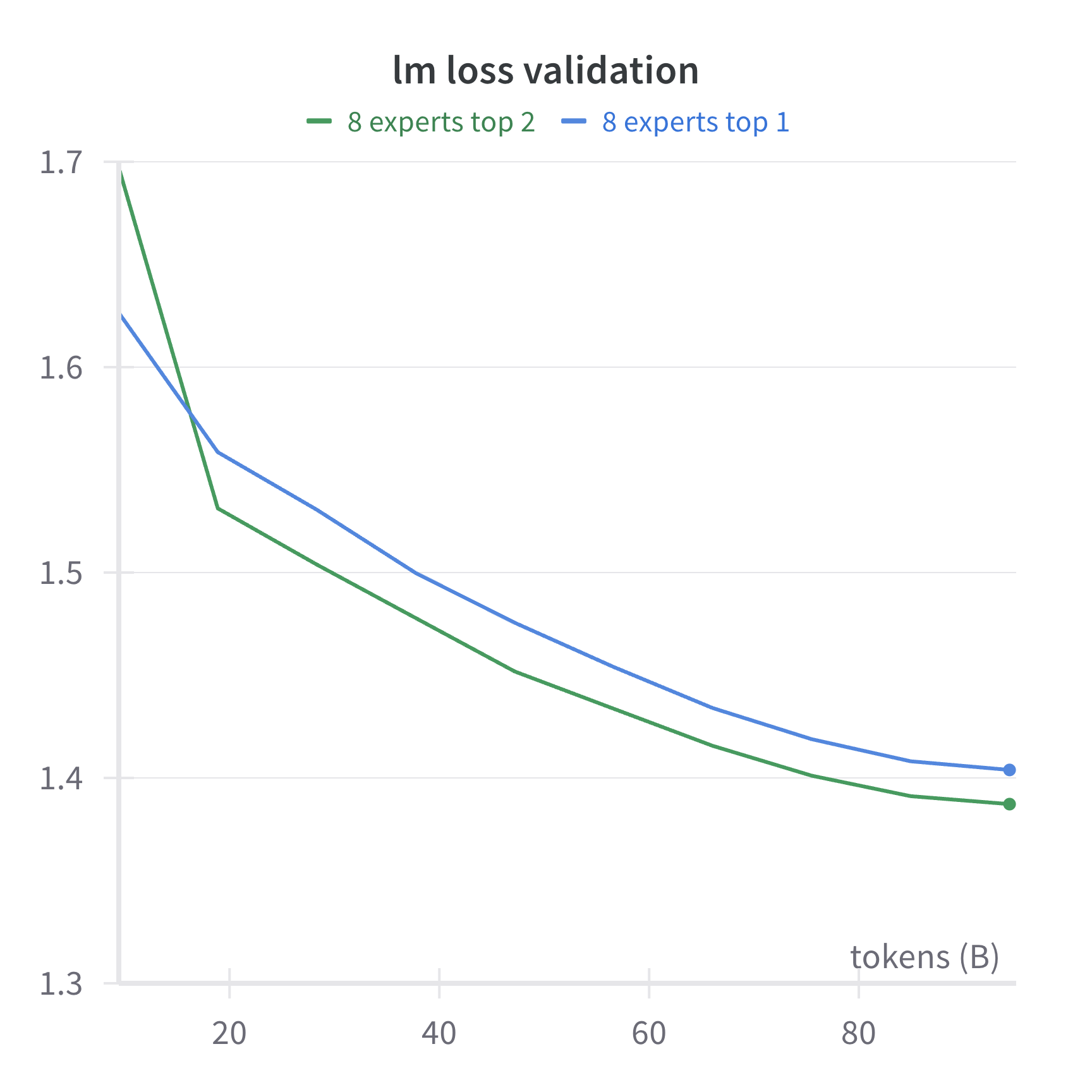}
        \caption{Upcycling Nemotron-4 15B}
    \end{subfigure}
    \caption{Increasing topK from 1 to 2 improves accuracy but requires extra compute}
    \label{fig:t1t2}
\end{figure}

Top-2 routing is often used with MoE models~\cite{jiang2024mixtral,zoph2022st}. While this increases the amount of compute required to run the model, it helps in achieving better accuracy. We compared increasing topK for both coarse-grained and fine-grained MoE upcycling. Figure~\ref{fig:t1t2} shows that 8 experts top-2 (E8G1T2) performed better than top-1 (E8G1T1) on both upcycling Nemotron 2B and Nemotron-4 15B. On 15B, the top-2 achieved lower training loss than top-1 ($1.35757$ vs $1.38031$). Previous works~\cite{zoph2022st} have shown that a tradeoff with wall clock time rather than compute is a better metric and in such cases topK greater than granularity might make more sense. This claim heavily depends on the actual implementation and we leave it as a future systems study.

\begin{figure}[!h]
    \centering
    \begin{subfigure}[b]{0.48\textwidth}
        \centering
        \includegraphics[width=1\linewidth]{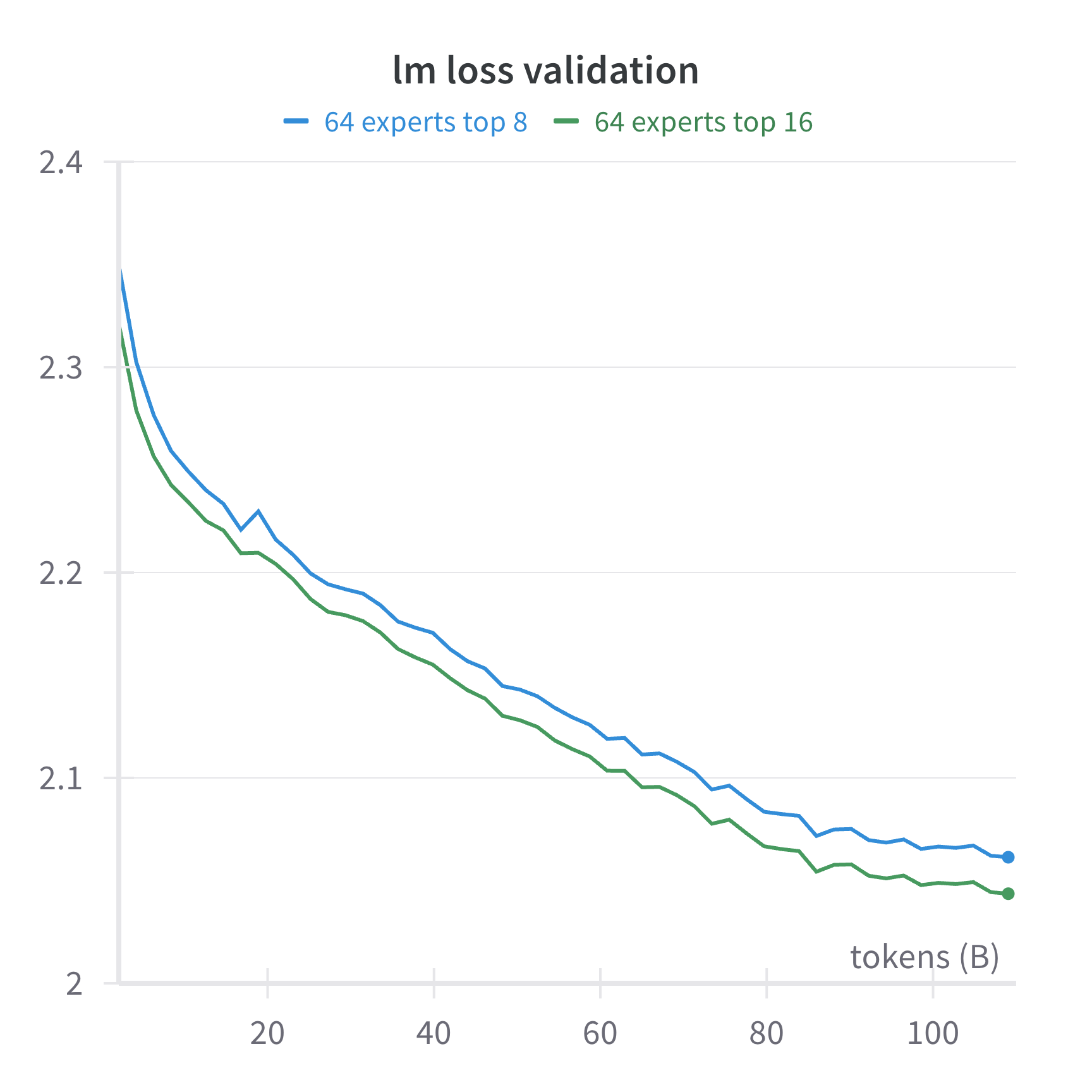}
        \caption{Upcycling Nemotron 2B}
    \end{subfigure}
    \hfill
    \begin{subfigure}[b]{0.48\textwidth}  
        \centering 
        \includegraphics[width=1\linewidth]{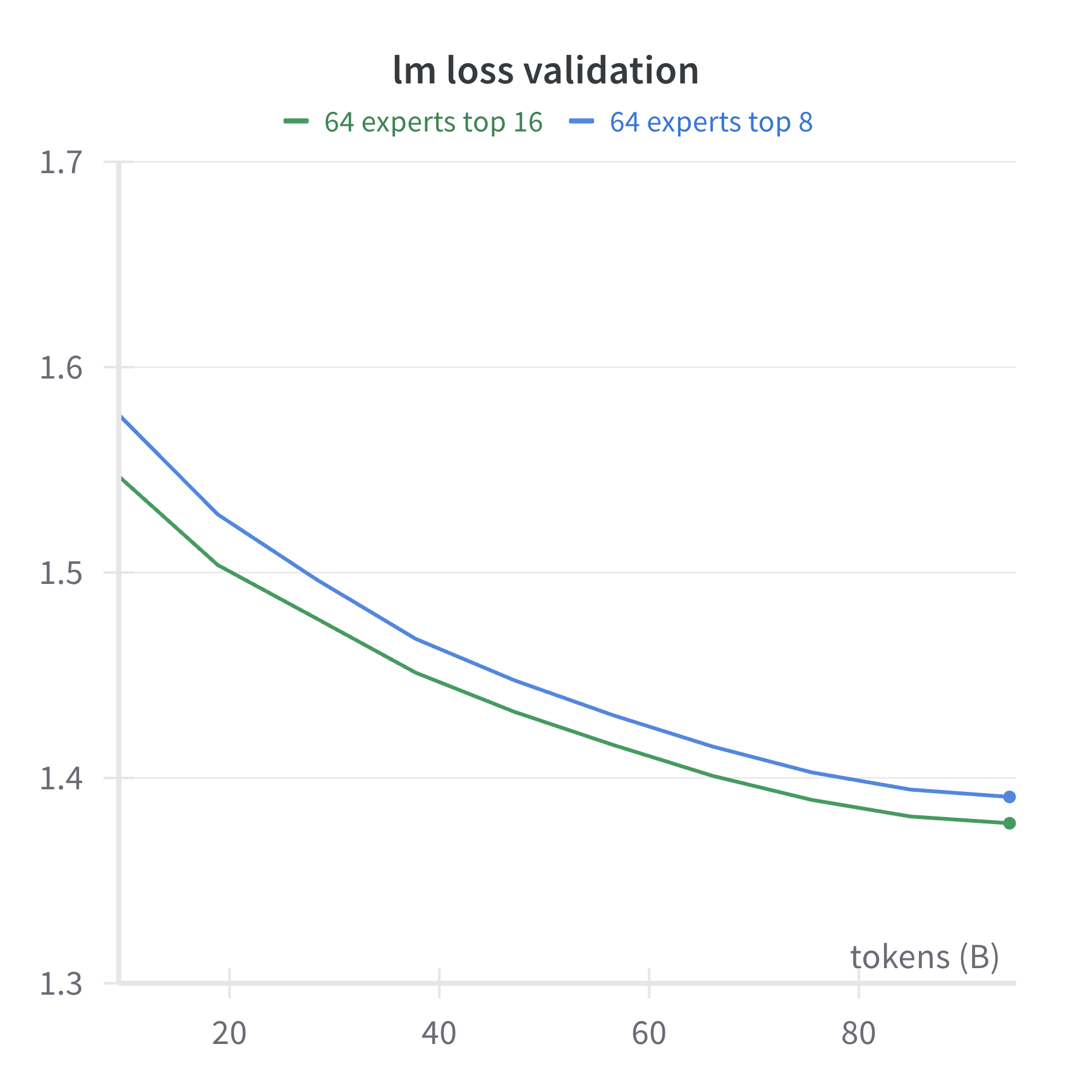}
        \caption{Upcycling Nemotron-4 15B}
    \end{subfigure}
    \caption{E8G8T8 vs. E8G8T16 (64 experts top-8 vs top-16, 1/8 expert intermediate size). Increasing TopK also helps granular models.}
    \label{fig:t8t16}
\end{figure}

\subsection{Promoting Expert Diversity: Weight Permutation and Reinitialization}
We experimented with the weight permutation and reinitialization methods proposed in Qwen 2~\cite{bai2023qwen}. Weight permutation permutes the FFN weights before copying it into each expert. Weights reinitialization randomly reinitializes 50\% of expert weights. In our experiments, we did not find any improvement with these two methods on upcycling Nemotron 2B. Due to limited compute, we did not experiment with these techniques on a larger network.

\subsection{Shared Experts}
\begin{figure}[!h]
    \centering
    \includegraphics[width=0.5\linewidth]{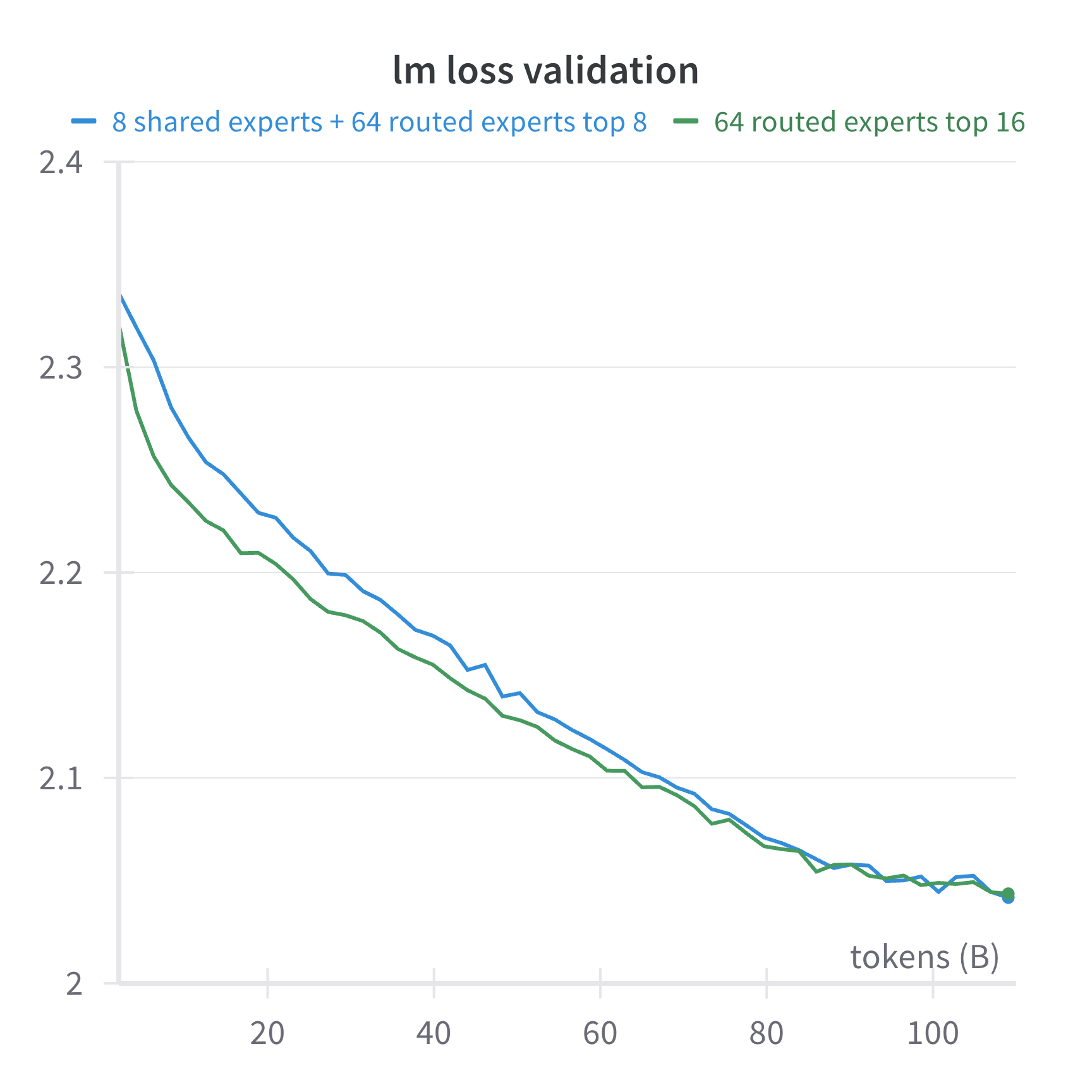}
    \caption{MoE with shared expert achieves similar accuracy as iso-FLOP fine-grained MoE in upcycling Nemotron 2B}
    \label{fig:sharedexperts}
\end{figure}
We experimented with the shared experts approach proposed in Deepseek-MoE~\cite{dai2024deepseekmoe}. Shared experts are always `on' i.e. every token is routed to the shared expert. They act like a dense layer in parallel with the MoE experts. As shown in Figure~\ref{fig:sharedexperts}, we compared 8 shared experts + 64 experts top-8 which is iso-FLOP with 64 experts top-16 on Nemotron 2B. We found that shared expert performed on par with the iso-FLOP, no shared expert counterpart. We did not switch to using a shared expert since we did not see any accuracy improvement by using it.

\section{Large Scale Upcycling}
To ensure our proposed recipes work on large scale (model size and token count) regimes, we compared upcycling against continued training Nemotron-4 15B base model on 1T tokens. We decided to test only 2 variants: (1) E8G8T8 and (2) E8G1T2, due to the high training compute requirement  (0.3 yottaFLOPs). We chose E8G8T8 with virtual group initialization since it worked better than E8G1T1 in our ablations. We chose E8G1T2 to show the effect of increase in compute and upcycling FLOPs. We chose E8G1T2 over E8G8T16 to show that our recommendations, including weight scaling along with softmax-then-topK router, work well for non-granular cases well. Non-granular use-cases are important to study since they practically achieve a better GPU FLOP utilization than their iso-FLOP granular counterparts. 

For 1T upcycling, we initialized lr to min-lr of pretraining ($4.5e$-4) and used a peak-lr of $3e$-4. We used cosine decay and decayed the lr to $1/100$-th of the pretraining min-lr as done for the original continued training of Nemotron-4 15B~\cite{parmar2024reusedontretrainrecipe}. We also used a higher batch size for the E8G8T8 model than E8G1T2 since E8G1T2 receives more tokens per expert on average. As shown in Table~\ref{tab:1t}, upcycled E8G8T8 (64 experts top-8 1/8 expert hidden size) achieved a \textbf{4.1\%} lower validation loss than the dense continued training model while being iso-FLOP. It also achieved a better MMLU score of 66.2 (vs 65.3 for dense). We observed that the percentage difference in MMLU is sensitive to the continuous training data and the difference increased along with the token count - so longer token horizons favor MoE models. With increased training FLOPs, we also upcycled E8G1T2 (8 experts top-2) which achieved \textbf{5.2\%} lower validation loss and an even better MMLU of 67.6. 

\begin{table}[!ht]
\centering
\begin{tabular}{l|c|c}
\hline
\textbf{Model} & \textbf{val loss} & \textbf{MMLU (5 shots)} \\ \hline
Nemotron-4 15B~\cite{parmar2024nemotron} &   1.623     & 59.3           \\ 
Nemotron-4 15B continued training~\cite{parmar2024reusedontretrainrecipe} & 1.377        & 65.3           \\ 
Nemotron-4 15B upcycling E8G8T8 & 1.320  & 66.2           \\
Nemotron-4 15B upcycling E8G1T2 & 1.306 & 67.6           \\ 
\end{tabular}
\caption{Upcycling Nemotron-4 15B on 1T tokens}
\label{tab:1t}
\end{table}

\section{Related Work}

\subsection{Mixture of Experts (MoE) Models}
Mixture of Experts (MoE) models~\cite{jacobs1991adaptive,shazeer2017outrageously} have gained significant attention in the field of large language models due to their ability to scale model capacity while maintaining computational efficiency. The MoE architecture employs a gating mechanism to selectively activate a subset of expert networks for each input token. This approach allows for an increased model capacity without a proportional increase in computational cost during inference and training.

Recent work has focused on improving the scalability and efficiency of MoE models. Switch Transformer~\cite{fedus2022switch} simplified the MoE architecture by using a top-1 routing mechanism and demonstrated the ability to scale to trillion-parameter models. The GShard framework~\cite{lepikhin2020gshard} addressed challenges in training large-scale MoE models, introducing techniques such as expert capacity thresholds and local group dispatching to improve load balancing and training stability.

Research has explored various aspects of expert specialization and routing mechanisms in MoE models~\cite{xue2024openmoe,muennighoff2024olmoe}. Studies have investigated the impact of the number of experts on model performance, finding that increasing the number of experts leads to improved sample efficiency and faster training, albeit with diminishing returns beyond certain thresholds~\cite{krajewski2024scaling}. The choice of routing algorithm (e.g., top-1 vs top-2) and gating function (e.g., softmax vs sigmoid) has also been examined~\cite{csordas2023approximating,chi2022representation,nguyen2024sigmoid}, with Mixtral 8x7B switching softmax and topK in the router~\cite{jiang2024mixtral}.

\subsection{Upcycling and Model Expansion}

The concept of upcycling in the context of MoE models refers to the practice of leveraging pre-trained dense models to initialize MoE architectures. This approach has gained traction as a way to efficiently create large-scale MoE models while benefiting from the knowledge captured in existing pre-trained checkpoints. Notable work in this area includes:

\noindent\textbf{Sparse Upcycling}: \cite{komatsuzaki2022sparse} proposed a method for training MoE models from dense checkpoints, demonstrating the ability to expand model capacity while maintaining or improving performance. Recently, Qwen~\cite{bai2023qwen} and Deepseek~\cite{bi2024deepseek} have adopted this approach. However, the recipe to scale upcycling beyond 1B parameters is not well known.

\noindent\textbf{Network Growth}: Research on model expansion techniques, such as those explored in the Gopher model~\cite{rae2021scaling}, has shown that it's possible to significantly increase model size while maintaining performance comparable to or better than models trained from scratch.

\noindent\textbf{Progressive Expansion}: Approaches like LLAMA PRO~\cite{wu2024llama} have investigated progressive expansion techniques, where model size is increased gradually during training~\cite{pan2023reusing,gesmundo2023composable,panigrahi2024efficient,wang2023learning}.

\subsection{Challenges and Ongoing Research}

Despite the promising results in upcycling and MoE pretraining, several challenges remain active areas of research:

\noindent\textbf{Expert Collapse}: The phenomenon of expert collapse, where certain experts become underutilized or inactive, has been observed in MoE training. While some studies suggest that expert collapse may not necessarily harm model accuracy, addressing this issue remains an important consideration in MoE design~\cite{fan2024towards,chi2022representation}.

\noindent\textbf{Load Balancing}: Ensuring an even distribution of work across experts continues to be a challenge, with various approaches proposed to improve load balancing, including auxiliary losses and specialized routing mechanisms. Skywork-MoE~\cite{wei2024skywork} used aux loss to promote expert diversity during upcycling. BTX~\cite{sukhbaatar2024branch,zhang2024bam} trained experts on different tasks and then mixed them. DBRX~\footnote{https://www.databricks.com/blog/introducing-dbrx-new-state-art-open-llm} added weight norm on experts' weights. 

\noindent\textbf{Efficient Training and Inference}: Ongoing work focuses on optimizing the training and inference processes for MoE models, including techniques for reducing communication costs and improving parallelization strategies. Deepseek-MoE~\cite{dai2024deepseekmoe} and Snowflake Arctic~\footnote{https://www.snowflake.com/blog/arctic-open-efficient-foundation-language-models-snowflake/} parallelized shared expert to increase utilization. MegaBlocks~\cite{gale2023megablocks} reformulated MoE computation in terms of block-sparse
operations. Scattermoe~\cite{tan2024scattered} optimized memory for finegrained MoE. 

\section{Conclusion}
In this work, we conducted an extensive study of upcycling techniques and best practices for billion-parameter scale language models. We proposed a ``virtual group" initialization scheme and weight scaling approach to successfully enable both coarse-grained and fine-grained MoE upcycling at scale. We found that upcycling outperforms continued dense model training for the same amount of compute on both 2B and 15B parameter models. Based on the target inference and available upcycling FLOPs, an architecture that uses more FLOPs like E8G1T2 can achieve better accuracy than dense iso-FLOP MoE models. On our 2B ablations, we found that upcycling needs a different set of hyper-parameters than fine-tuning. Softmax-then-topK expert routing performs better than topK-then-softmax in the MoE router during upcycling. Higher granularity MoEs boost upcycling accuracy but require a more careful weight scaling and sharding strategy and also lead to a lower GPU FLOP utilization. In a purely FLOP-bound scenario, using virtual group init with granular model upcycling seems to be the best strategy. 
We hope this work illuminates the details of upcycling billion-parameter scale MoE models. Future directions include studying upcycling in larger models, improving expert diversity and utilization, and co-optimizing model architectures and system designs.

\section{Acknowledgement}
We would like to thank Aditya Vavre, Deepak Narayanan, Dennis Liu, Helen Ngo, Jakub Krajewski, June Yang, Jupinder Parmar, Marcin Chochowski, Mostofa Patwary, Ray Wang, Xin Yao, and Zeeshan Patel for their helpful contributions and discussions.

\bibliographystyle{unsrt}
\bibliography{references}

\appendix

\section{Virtual Grouped Router}\label{sec:virtual_grouping_code}
The following pseudo code illustrates our virtual grouped router approach for upcycling into E2G2T2:

\begin{lstlisting}[frame=single]
# sharding FFN into expert shards G=2
FFN     =   [FFN_0, FFN_1]

# copying to form multiple experts; E=2
experts = [FFN_0, FFN_1,
           FFN_0, FFN_1]

# random initialized router
router_prob = tensor([0.4, 0.2,
                      0.3, 0.1])
router_top2 = tensor([0.4, 0.0,
                      0.3, 0.0])

# For dense model
FFN(x) = FFN_0(x) + FFN_1(x)

# In case of MoE
FFN_moe(x) = router_top2@experts = 0.4 FFN_0(x) + 0.3 FFN_0(x) $\ne$FFN(x)

# virtual group - initialize every group with same weights
router_prob = tensor([0.3, 0.3,
                      0.2, 0.2])
router_top2 = tensor([0.3, 0.3,
                      0.0, 0.0])
# one of each FFN shard is guaranteed to be selected
FFN_moe(x) = 0.3 FFN_0 + 0.3 FFN_1 $\approx$ FFN(x) / 4
\end{lstlisting}

\end{document}